# Degeneracy Counting Quantum Algorithm using Decoherence

Malay Das [1], Mark A. Novotny[2] and Yaroslav Koshka[1]

[1] Department of Electrical and Computer Engineering, Mississippi State University, Mississippi State, MS 39762 USA.
[2] Department of Physics and Astronomy, Mississippi State University, Mississippi State, MS 39762 USA.
**E-mail:** mmd402@msstate.edu , man40@msstate.edu , ykoshka@ece.msstate.edu

***Abstract*—Counting the global optima of a classical optimization problem is a #P-hard task. We develop the canonical thermal pure quantum (CTPQ) state-based degeneracy counting (CTPQsd#) algorithm that determines the number of global optima of a classical optimization problem P by measuring only a small probe S, without finding individual minima. The method exploits a perturbative relation between the decoherence measure of S and the degeneracy of P when S and P are together in a CTPQ state. We provide the first numerical demonstration that this relation can be used to count the global minima, applying it to problems encoded by diagonal random-energy Hamiltonians as a maximally unstructured testbed for classical binary optimization problems. Classical simulations of up to 20 problem qubits quantify the algorithm's sensitivity to variations in the temperature of the CTPQ state, the Hamiltonian energy range, the problem size, and degeneracy. We establish the temperature threshold for determining the exact degeneracy and identify a second, lower threshold that provides a temperature window to count near-degenerate minima within a user-defined energy tolerance. By confining measurement to S, the protocol replaces tomography over the exponentially large problem Hilbert space with tomography over a small probe represented by only four qubits.**

## I. Introduction

In many practical systems, the landscape of a target function, whether an objective function, a loss function, or an energy landscape, often exhibits multiple global extrema, a condition usually referred to as exact degeneracy [1]. In the context of physics and quantum computing, the count of these global extrema is referred to as the ground-state (GS) degeneracy of the system's Hamiltonian, into which the target cost function is usually encoded, e.g., a Quadratic Unconstrained Binary Optimization (QUBO) [2], an Ising model [3], or a higher-order many-body Hamiltonian [4]. Furthermore, even when a system possesses a unique true global extremum, it frequently harbors numerous local extrema whose values lie within a marginally small tolerance of the global optimum. This phenomenon, recognized as quasi-degeneracy [5] or approximate degeneracy [6], is highly relevant across disciplines. In physics, it characterizes low-lying excited states separated by a minimal energy gap [2], while in optimization, it represents a set of near-optimal or good-enough solutions [7,8].

Often, especially in complex or large-scale problems, finding multiple solutions associated with a cost function is computationally expensive [9], infeasible, or unnecessary. Instead, knowing the exact or estimated count of global or near-optimal solutions, i.e., the exact degeneracy or approximate degeneracy, may be a useful computational goal. In physical systems described by QUBO or Ising Hamiltonians, the ground-state degeneracy directly determines the residual entropy, a key thermodynamic quantity characterizing frustrated and disordered phases [10,11]. An absence of solutions to a constraint satisfaction problem (CSP) indicates an over-constrained system [12], while infinitely many solutions indicate an ill-posed problem that may require reformulation or regularization [13]. Existence of only one solution, i.e., solution uniqueness, is one of the necessary conditions for a well-posed problem [14] with a definitive answer. Conversely, a high count of exact or near-optimal solutions suggests non-uniqueness and redundancy. This can be advantageous, enabling exploration of diverse alternatives and trade-offs, or problematic, as selecting among degenerate solutions requires additional criteria beyond the objective value alone [15]. In summary, efficiently determining the system's degeneracy (i.e., the number of GSs/global extrema) without actually finding all GSs may be important for many applications.

In computational complexity theory, exactly counting the number of solutions to a constraint problem belongs to the #P complexity class [16]. In most practical counting problems [17,18], exact computation is intractable under standard complexity-theoretic assumptions. Consequently, much of the theoretical and practical focus has shifted to approximate counting algorithms. However, these methods purposefully trade absolute precision for scalability, and their applicability is based on structural assumptions that are not always valid.

To mitigate the inherent issues of classical counting algorithms, quantum computing offers alternative approaches, such as Quantum Amplitude Estimation (QAE) [19]. However, these quantum counting algorithms require fully coherent execution of deep circuits on a Quantum Processing Unit (QPU). Furthermore, hardware requirements render these methods impractical for current Noisy Intermediate-Scale Quantum (NISQ) hardware [20].

Consequently, the field has pivoted toward hybrid quantum-classical algorithms and variational algorithms. Examples include QAE variants, such as Iterative QAE (IQAE) [21] and Maximum Likelihood QAE (MLQAE) [22], and the Variational Quantum Counting algorithm, such as VQCount [23], employing a Quantum Alternating Operator Ansatz (QAOA) within the Jerrum–Valiant–Vazirani self-reducibility framework [24]. Hybridization also enhances classical Markov Chain Monte Carlo (MCMC) algorithms [25]. Further, there exist quantum-inspired classical methods based on tensor network contractions [26,27].

Despite these advancements, the practical utility of current hybrid algorithms remains severely constrained by several fundamental bottlenecks. First, variational methods are susceptible to the barren plateau phenomenon [28,29], rendering gradient-based optimization of the cost function exponentially hard and practically infeasible at scale. Classical optimization required to navigate the landscape of the variational algorithm's cost function is, in general, NP-hard [30]. Furthermore, reconstructing the full density matrix of the $n$-qubit system, a task whose shot complexity scales as $O\left(\frac{4^n}{\epsilon^2}\right)$ to achieve trace-distance precision $\varepsilon$, is an exponential overhead that renders full-register tomography intractable beyond small system sizes [31,32]. Concurrently, on noisy hardware, readout errors on individual qubits accumulate multiplicatively across the measured register, rendering measurements unreliable as system size grows [33].

In contrast to the classical methods mentioned earlier in this section, some hybrid quantum–classical algorithms can determine or estimate the number of solutions without enumerating them (i.e., without finding all solutions) [21,22,23], which makes them more relevant to the objectives of this work. However, the applicability of these methods to the specific counting problem targeted in our work, namely, determining the number of existing GSs without enumerating all of them, is yet to be explicitly explored.

To address the measurement bottleneck [34,35] and other aforementioned limitations, this paper investigates a method based on decoherence of a (small) probe system interacting with the system under investigation, as proposed by Novotny et al. [36,37]. In that prior work, measures of decoherence and thermalization of a canonical thermal pure quantum (CTPQ) state [38] at infinite [39] and finite [36] temperatures have been investigated. In general, decoherence is expected to drive the off-diagonal elements of the reduced density matrix of the system $S$ (hereafter referred to as the Probe system $S$) to zero in the eigenbasis of its Hamiltonian $H_S$. It was established that the decoherence measure $\sigma_S$ and the thermalization measure $\delta_s$ of $S$ remain finite (i.e., do not become zero) when a finite-dimensional quantum environment E (which in this work is employed to represent the Problem and hereafter is denoted as $P$) is in thermal equilibrium with it at finite temperature. Using perturbation theory with respect to the coupling strength $\lambda$ between the probe and environment (problem), Novotny and co-authors derived closed-form expressions [36] relating the degeneracy of the GS of a larger quantum Hamiltonian $H_P$ to decoherence (and also thermalization) measures of the smaller quantum probe $S$, where $S+P$ together are in a CTPQ state. A key prediction emerging from this analysis is that in the low-temperature limit, for weak or zero system-environment coupling and for a system $S$ with a degenerate ground state $g_S > 1$, the expectation value $E(\sigma_S^2)$ depends explicitly on the GS degeneracy $g_P$ of the environment (i.e., the problem) Hamiltonian $H_P$ [36]. A parallel result holds for the thermalization measure $\delta_S$, although the expression involves additional terms related to the specific heat and effective temperature of $S$ [36]. This result implies that the degeneracy $g_P$ of a potentially large problem system $P$ could be determined by reconstructing the reduced density matrix of a very small probe system $S$ (e.g., $N_S = 4$ qubits) via quantum state tomography, thereby reducing the measurement overhead from scaling with the full problem Hilbert space dimension, which could require thousands or tens of thousands of qubits, to scaling with only the probe dimension, which needs to be only a few qubits. Because this algorithm does not prescribe a specific thermalization mechanism, requiring only that the entirety $S+P$ be prepared in a CTPQ state, it is inherently modular and can be integrated as a localized probe measurement technique with a compatible state preparation method, e.g., Variational Quantum Thermalizer (VQT) [40] or Variational Quantum Imaginary Time Evolution (VarQITE) [41]. This offers a potentially scalable approach to determine exact or approximate degeneracy in complex systems using quantum computers.

Before the quantum degeneracy counting algorithm can be implemented on physical quantum hardware, it is essential to establish a proof of concept and characterize how the decoherence measure depends on key parameters of the algorithm. Although previous studies [36,37,42] derived analytical expressions for the decoherence measure in terms of the GS degeneracy $g_P$ and verified them numerically for a few specific spin-chain configurations, they did not demonstrate the feasibility of the approach nor develop a systematic procedure for using these expressions to extract $g_P$. While those earlier studies provided foundational analytical and numerical results, they were limited to a narrow class of one-dimensional spin-1/2 Heisenberg Hamiltonians with a ring topology, optionally augmented with additional couplings between non-neighboring environment spins or between system and environment spins. It remains to be established whether the degeneracy counting method extends to a broader range of problems described by very different Hamiltonians, which would support its generality. Further, it remained unclear whether the relationship between the degeneracy of interest and the decoherence measures in prior work [36] holds for classical diagonal Hamiltonians and is not limited to certain quantum Heisenberg Hamiltonians with nonzero off-diagonal elements. Finally, while the prior work identified key conditions for the validity of the algorithm, such as the restriction on the degeneracy of the probe system ($g_S > 1$), and the requirement of low temperature and weak or zero system–environment (or probe-problem) coupling, a systematic investigation of how sensitive the degeneracy extraction is to various parameters, and their practical bounds, has not been carried out.

This work investigates the capability of the quantum counting algorithm to determine the number of global minima in a sufficiently arbitrary classical binary problem encoded as a diagonal Hamiltonian $H_P$ with i.i.d. uniformly distributed diagonal energies, related to the classical (zero-transverse-field) limit of the quantum random energy model (QREM) [43,44] ($\Gamma = 0$). This is accomplished through numerical simulations of the composite system $S + P$ of a sufficiently small size on a classical computer. This work focuses on problems that can be encoded purely in the diagonal elements of a Hamiltonian, the category to which all classical optimization problems belong, and uses a random-energy (REM-like) Hamiltonian as a maximally unstructured representative of that class. The robustness of the counting algorithm is evaluated by varying different parameters such as the energy scale in the Hamiltonian represented as a uniform distribution in $[-R, R]$, the number of qubits in the problem system ($N_P$), degeneracy of the problem Hamiltonian ($g_P$), and temperatures ($T$) of the CTPQ $S + P$ state preparation, allowing for the characterization of their impact on finding $g_P$. The focus on classical (i.e., diagonal) Hamiltonians also allowed maximizing the $N_P$ (i.e., the number of qubits representing the binary variables of the problem under investigation) to 20 qubits that could still be simulated on a classical computer, since the absence of off-diagonal terms in $H_P$ reduces the computational and memory overhead required on the computer. Additionally, this work addresses a question of significant practical importance: the behavior of the algorithm under quasi-degeneracy [5,6], where multiple low-energy states have energy values close to, but not identical to, the GS energy $E_0$. Specifically, we investigate whether the method can still accurately determine the number of such near-degenerate minima and how the energy gap $\Delta g_P$ between the lowest-lying states affects the accuracy of the extracted count, a regime not explored in any prior theoretical studies.

This paper is organized as follows. Section II reviews the theoretical background, including the CTPQ, the measure of decoherence ($\sigma_S$), and summarizes the analytical relationships between this measure and the GS degeneracies of the system and environment derived in prior work [36]. Section III details the Hamiltonian models, computational methods employed, and quantum system configurations simulated in this paper. Section IV presents result on determining the GS degeneracy from decoherence measurements and how this determination depends on system size, temperature, and the degeneracy structure of the problem Hamiltonian. Section V concludes with a summary and a discussion of open questions, and the steps required to implement this algorithm on future quantum computing hardware.

## II. THEORY

### *A. Probe and problem systems - a system interacting with the environment*

Decoherence is a foundational concept in the theory of open quantum systems and plays a central role in the algorithm investigated in this work. Decoherence is traditionally viewed as an obstacle in quantum computing (QC), since it degrades coherent quantum superpositions into classical statistical mixtures (i.e., mixtures of classical probabilities), suppressing the off-diagonal elements of the density matrix in the pointer basis. However, a number of QC algorithms, including the approach explored in this work, treat decoherence as a resource rather than a liability [36,45,46]. Prior work [36,39] investigated decoherence of a quantum system $S$, which plays the role of the probe in the present framework, coupled to a quantum environment, here termed the problem system $P$. The entirety $S + P$ is a closed quantum system governed by the time-dependent Schrödinger equation (TDSE). The Hamiltonian of the entirety, using the notations adopted in this work, can be decomposed as

$$H = H_S + H_P + \lambda H_{SP} \quad (1)$$

where $H_S$, $H_P$, and $H_{SP}$ denote the probe $S$, problem $P$, and probe-problem interaction Hamiltonians, respectively, and $\lambda$ is the global coupling strength.

Within this framework, the state of the closed entirety composed of $N$ spin-1/2 particles is described by the state vector $|\psi(t)\rangle$ residing in the Hilbert space of dimension $D = 2^N$. $D_S = 2^{N_S}$ and $D_P = 2^{N_P}$ are dimensions of $S$ and $P$, respectively. The time evolution of this state is governed by the TDSE, and in natural units ($\hbar = 1$), is given by

$$|\psi(t)\rangle = e^{-itH} |\psi(0)\rangle = \sum_{i=1}^{D_S} \sum_{p=1}^{D_P} c(i,p,t) |i,p\rangle \quad (2)$$

where $\{|i,p\rangle\}$ form a complete orthonormal basis for the entirety in some chosen basis, with $|i\rangle$ and $|p\rangle$ denoting states of $S$ and $P$, respectively, and $c(i,p,t)$ are the expansion coefficients.

System $S$ is part of a larger closed system $S + P$, and the reduced density matrix of $S$ is given by

$$\begin{aligned} \rho_S(t) &\equiv Tr_P \rho_{S+P}(t) \\ &= Tr_P \sum_{p=1}^{D_P} \sum_{q=1}^{D_P} c^*(i,q,t) c(j,p,t) \\ &\quad |j,p\rangle\langle i,q| \\ &= Tr_P \sum_{p=1}^{D_P} c^*(i,p,t) c(j,p,t) |j\rangle\langle i| \end{aligned} \quad (3)$$

obtained by performing a partial trace over the degrees of freedom of the environment [36]. In equation (3), $\rho_{S+P}(t)$ is the density matrix of the entirety.

### *B. Decoherence measure*

The diagonal elements of the reduced density matrix $\rho_S(t)$ represent the occupation probabilities of the basis states, while the off-diagonal elements encode quantum coherences between the basis states. Decoherence manifests as the decay of phase coherence of the components in $S$ in the pointer basis arising from its interaction with the environment, i.e., our problem system $P$, effectively driving the system toward a classical statistical mixture. In the present work, the CTPQsd# (CTPQ state-based degeneracy counting) algorithm requires that the probe and problem are uncoupled ($\lambda = 0$). Because of that, the energy eigenbasis of $H_S$ serves as the natural pointer basis for the decoherence analysis. For the scope of the classical optimization problems targeted in this work, the basis of the eigenstates of $H_S$ and $H_P$ is also the computational basis, which corresponds to the states of the classical optimization (or degeneracy counting) problem. The degree of decoherence of the system $S$ is quantified by the measure $\sigma_S(t)$, defined as the square root of the sum of the squared moduli of the off-diagonal elements of the reduced density matrix in the eigenbasis of $H_S$, written as

$$\sigma_S(t) = \sqrt{\sum_{i=1}^{D_S-1} \sum_{j=i+1}^{D_S} |\tilde{\rho}_{ij}(t)|^2} \tag{4}$$

If $\sigma_S(t) = 0$, the system $S$ is in a state of full decoherence relative to the energy eigenbasis of $H_S$.

### *C. Quantum dynamics*

A unitary evolution of an isolated system governed by the TDSE cannot produce the irreversible decay of off-diagonal coherences in the density matrix $\rho(t)$ required for decoherence. To achieve decoherence, interaction with an environment (e.g., the problem system $P$ in our case) is necessary. This process drives $S$ towards a thermal steady state, essential for the CTPQsd# algorithm. The necessary interaction between probe $S$ and problem $P$ is quantified by a non-zero coupling strength $\lambda$ in equation (1) [36,39]. When $S$ and $P$ are initialized in a product state, a nonzero $\lambda$ is essential for the dynamics to generate entanglement and drive the system toward a thermal steady state. The dynamics of a specific Hamiltonian $H$ that drives the system from a physically prepared initial state towards one with the statistical properties of a canonical thermal steady state is a distinct and nontrivial question that depends on the details of the Hamiltonian, the geometry, and the nature of the couplings [39]. A canonical thermal steady state at a specific temperature can be achieved during simulation for a Hamiltonian with zero $\lambda$ by applying the imaginary-time projection to an infinite-temperature state (see Section II-E).

This canonical steady state, in which thermal expectation values are reproduced at the level of individual pure states, is referred to as a scaling state [39] or a canonical thermal state [36] in prior literature. Following Sugiura and Shimizu [38], we refer to this canonical thermal state as the Canonical Thermal Pure Quantum (CTPQ) state.

### *D. Canonical thermal pure quantum state*

In quantum statistical mechanics, the thermal steady states are conventionally represented as mixed quantum states within the ensemble formulation [47]. An alternative pure-state framework based on CTPQ states [48] demonstrates that a single pure quantum state can reproduce canonical ensemble predictions.

A state $\mid \psi\rangle$ is a CTPQ state if

$$\langle A\rangle_N^{\psi} \overset{P}{\rightarrow} \langle A\rangle_N^{ens} \tag{5}$$

uniformly for every operator A in the thermodynamic limit $N \to \infty$. Here $\langle A\rangle_N^{\psi} \equiv \langle\psi \mid A \mid \psi\rangle / \langle\psi \mid \psi\rangle$, $\langle A\rangle_N^{ens}$is the ensemble average, and $\overset{P}{\rightarrow}$ denotes convergence in probability [38]. That is, for an arbitrary $\epsilon > 0$, there exists a function $\eta_\epsilon(N) \to 0$ that vanishes in the thermodynamic limit and satisfies

$$P(| \langle A\rangle_N^{\psi} - \langle A\rangle_N^{ens} | \geq \epsilon) \leq \eta_\epsilon(N) \tag{6}$$

where$P(x)$ denotes the probability of event $x$. This means that for sufficiently large $N$, a single realization of a CTPQ state can reproduce the thermal steady state expectation value of any observable, up to statistical or quantum fluctuations [38].

At infinite temperature, all energy eigenstates are equally weighted, so the CTPQ state reduces to a random state $\mid \psi_0\rangle$ drawn uniformly from the $D$-dimensional Hilbert space of the entirety as

$$\mid \psi_0\rangle = \sum_{k=1}^{D} d_k \mid E_k\rangle \tag{7}$$

where $\{d_k\}$ are random Gaussian coefficients normalized to unity $\sum_{k=1}^{D} d_k^* d_k = 1$. The Gaussian random numbers $d_k$are generated using the Box-Muller method

$$d_k = \frac{c_k' + i b_k'}{\sqrt{\sum_{k'=1}^{D}[(c_{k'}')^2 + (b_{k'}')^2]}} \tag{8}$$

where

$$c_k' = \sqrt{-2ln(r_0^{(k)})}cos(2\pi r_1^{(k)}) \tag{9}$$

and

$$b_k' = \sqrt{-2ln(r_0^{(k)})}sin(2\pi r_1^{(k)}) \tag{10}$$

where $r_0^{(k)}$and $r_1^{(k)}$are independent random numbers distributed uniformly on $[0,1)$. Mathematically, this state is equivalent to a point sampled from the uniform distribution on the hypersphere in Hilbert space [49].

### *E. Preparation of a finite temperature CTPQ state*

On a quantum computer, preparing a CTPQ state of the entirety from an initial product state would require real-time evolution or an equivalent state-preparation protocol. In the present work, we instead construct a CTPQ state of the entirety at finite temperature $T$ on a classical computer by applying the imaginary-time projection operator $e^{-\beta H/2}$ to an infinite temperature CTPQ state $\mid \psi_0\rangle$ followed by normalization as

$$\mid \psi_\beta\rangle = \frac{e^{-\beta H/2} \mid \psi_0\rangle}{[\langle\psi_0 \mid e^{-\beta H} \mid \psi_0\rangle]^{1/2}} \tag{11}$$

where $\beta = 1/k_B T$ [36]. This construction is justified by the fact that for any quantum observable $A$ of the entirety $S+P$, the expectation value satisfies

$$\langle \psi_\beta \mid A \mid \psi_\beta \rangle \approx \frac{TrAe^{-\beta H}}{Tre^{-\beta H}} \tag{12}$$

in the thermodynamic limit, recovering the canonical ensemble average. In the eigenenergy basis $\{\mid E_k\rangle\}$ of the Hamiltonian $H$ of the entirety, the state is given by

$$\mid \psi_\beta \rangle = \sum_{k=1}^{D} \frac{d_k e^{-\beta E_k/2}}{\sqrt{\sum_{k'=1}' e^{-\beta E_k'}}} \mid E_k \rangle = \sum_{k=1}^{D} a_k \mid E_k \rangle \tag{13}$$

where the coefficients $a_k$ incorporate both the Boltzmann weighting $e^{-\beta E_k/2}$ and the random amplitudes $d_k$ of the initial state [36] as

$$a_k = \frac{d_k p_k^{1/2}}{\sqrt{\sum_{k'=1}^{D} \mid d_{k'} \mid^2 p_{k'}}} \tag{14}$$

$$p_k = \frac{e^{-\beta E_k}}{\sum_{k'}' e^{-\beta E_{k'}}} \tag{15}$$

In general, the probability density of the coefficient $a_k$ is not Gaussian, unlike $d_k$ at infinite temperature, because normalized Boltzmann probabilities $p_k$ introduce correlations among the coefficients. In the infinite-temperature limit ($\beta \to 0$), the Boltzmann weights become uniform and $\mid \psi_\beta\rangle$ reduces to $\mid \psi_0\rangle$. As the temperature decreases, the Boltzmann weighting biases the state toward low-energy configurations,

A notable property of the CTPQ state is that, even in the absence of coupling between $S$ and $P$ parts of the Hamiltonian (i.e., $\lambda = 0$), the CTPQ state of the entirety for such an uncoupled Hamiltonian is generally entangled and cannot be written as a product state $\mid \psi_S\rangle \otimes \mid \psi_P\rangle$. This corresponds to the fact that the random coefficients $d_{i,p}$ in the joint eigenbasis are not factorizable as $d_i d_p$.

The development of efficient quantum circuits for CTPQ state preparation is beyond the scope of the present work. We therefore assume that the entirety $S+P$ has reached a CTPQ state and focus on the algorithmic and analytical consequences of this assumption. For numerical validation, we create the CTPQ state classically via imaginary-time projection, which circumvents the need for explicit time evolution but limits the problem system size to 20 qubits due to the exponential growth of the Hilbert space dimension.

## F. Relevant results of the perturbation theory analysis

The steady state of the entirety $S+P$, represented by a CTPQ state, possesses well-defined statistical properties that enable analytical derivation of the scaling relations for decoherence measures [36]. The CTPQ states previously studied have yielded predictions for decoherence measures both at infinite temperature [39] and at finite temperatures [36].

At infinite temperature, the decoherence measure obeys the scaling relation

$$\sigma_S \propto \frac{1}{\sqrt{D_P}} \tag{16}$$

provided the state of the entirety is a CTPQ state [39]. However, this expression does not include the GS degeneracy count, which is precisely the quantity targeted by the method analyzed in this work. This motivates the analysis of CTPQ states at finite temperatures. When the entirety $S+P$ is prepared in a CTPQ state, perturbation theory in the probe–problem coupling strength $\lambda$ provides analytical expressions for the decoherence measure [36]. The expectation value of $\sigma^2$ of an uncoupled entirety ($\lambda = 0$) in the entangled CTPQ state is

$$E(\sigma_S^2) = [\frac{D}{(2(D+1))}][1 - e^{-2\beta(F_S(2\beta)-F_S(\beta))}](e^{-2\beta(F_P(2\beta)-F_P(\beta))}) \tag{17}$$

where $F_P(\beta) = -\beta^{-1} ln\, Z_P(\beta)$ is the free energy of $P$.

These perturbation results [36,37] hold for arbitrary Hamiltonians $H_S$ and $H_P$, providing a general framework for predicting the decoherence behavior of any quantum probe $S$ weakly coupled, or uncoupled, to a quantum environment $P$ representing the problem in our work. The exact expressions for the expected decoherence values can be simplified at two extreme temperature limits as

$$lim_{\beta\to 0}\, E(\sigma_S^2) = \frac{D_S - 1}{2(D_S D_P + 1)} \tag{18}$$

$$lim_{\beta\to \infty}\, E(\sigma_S^2) = \frac{g_S - 1}{2 g_S g_P}[1 - \frac{D_S D_P}{(D_S D_P + 1) g_S g_P}] \tag{19}$$

where $g_S$ and $g_P$ denote the GS degeneracies of $S$ and $P$, respectively. It has been shown that the above observations are still valid for a small spin environment, even when the probe and problem are of comparable sizes [42].

Remarkably, the low-temperature expression implies that, when $g_S > 1$, $\sigma_S > 0$, and measurements performed solely on the probe $S$ can, in principle, reveal the GS degeneracy $g_P$ of the problem system $P$. This connection between the decoherence properties of $S$ and the spectral structure of $P$ forms the theoretical foundation of the algorithm explored in this work.

## G. Random-energy-model background

The Hamiltonians examined in the present study are structurally distinct from Heisenberg-type spin systems with the ring topology of prior work [36,39]. One of the objectives of the present work is to demonstrate that the perturbation-theoretic predictions of equations (25)-(28) of Novotny et al.'s work [36] can be used to determine the degeneracy of problems described by a structurally different class of Hamiltonians, exemplified here by a diagonal, REM-like Hamiltonian related to the quantum random energy model (QREM). QREMs [43] are non-Heisenberg-type Hamiltonians and can be described as a class of mean-field spin glass models with a $p$-spin interaction defined on $N$ Ising-type spins

$$s = (s_1, \ldots, s_N) \in \{-1,1\}^N \tag{20}$$

where $s_1$, $s_N$ denotes individual qubit spins. For fixed $p \in [1, \infty)$ the interaction energy of these spins is given by

$$H_p(s) = \frac{1}{N^{\frac{p-1}{2}}} \sum_{j_1,\dots j_p=1}^{N} g_{j_1 \dots j_p} s_{j_1}, \dots, s_{j_p} \tag{21}$$

where $g_{j_1,\dots j_p}$ are independent and identically distributed (i.i.d.) Gaussian random variables with unit variance. The case $p = 2$ corresponds to the Sherrington-Kirkpatrick model [50], while the limit $p \to \infty$ yields Derrida's Random Energy Model (REM) [44, 51]. In the REM limit, the spin–spin correlations vanish, and the $2^N$ energy values of $H_\infty(s)$ form an i.i.d. Gaussian process on the hypercube of spins $\{-1,1\}^N$.

While the Quantum Random Energy Model (QREM) typically introduces non-diagonal elements via a transverse field, our framework restricts the system to a purely classical landscape. The Ising spin configurations are identified with the eigenstates of the $z$-components of $N$ spin-1/2 operators, and the classical energy function $H_p$ is promoted to a diagonal Hamiltonian matrix $H$ acting on the corresponding $2^N$-dimensional Hilbert space at zero transverse field.

This work, which focuses on classical optimization problems and their degeneracy, utilizes REM-like (diagonal random-energy) Hamiltonians whose levels are drawn from a uniform rather than a Gaussian distribution of the canonical REM. The Hamiltonian matrix is diagonal in the computational basis, while quantum spins encode binary classical variables, with the two spin states corresponding to the two possible binary values.

## III. Method

### *A. Overview of the CTPQsd# algorithm*

For a given problem size $N_P$ and problem degeneracy $g_P$, the CTPQsd# algorithm proceeds in four steps:

1. Construct a probe Hamiltonian $H_S$ with a GS degeneracy $g_S > 1$, so that the low-temperature limit of equation (19) provides a nonzero $\sigma_S$. For the reasons discussed below, we use a diagonal $H_S$ in the present work.

2. Construct a problem Hamiltonian $H_P$, whose GS degeneracy $g_P$ represents the quantity of interest to be determined by the algorithm. Since this work focuses on determining $g_P$ of classical problems, only classical (i.e., diagonal) $H_P$ are investigated.

3. Prepare the uncoupled entirety $S + P$ ($\lambda = 0$) in a finite-temperature CTPQ state $|\psi_\beta\rangle$. In this work, it is achieved by imaginary-time projection of a random infinite-temperature CTPQ state $|\psi_0\rangle$ as described in Sections II-D and II-E.

4. Trace out $P$ to obtain $\rho_S$, compute $\sigma_S$ via equation (4). For larger systems, as in [36], a single such measurement may suffice. Nevertheless, one can average $\sigma_S^2$ over $n$ independent CTPQ realizations and recover the estimate of $g_{P,Meas}$ from the low-temperature limit of equation (19). To achieve a statistically robust estimate of the problem degeneracy, it is necessary to compute the expectation value $E(\sigma_S^2)$ by averaging over an ensemble of independent CTPQ state realizations.

Throughout this work, the probe and problem are uncoupled ($\lambda = 0$). In addition to being consistent with the regime of validity of equation (19), this design choice yields two operational advantages. First, the energy eigenbasis of $H_S$ becomes the computational basis of the probe $S$, eliminating the need for an additional basis transformation before computing $\sigma_S$. Second, the entirety Hamiltonian $H = H_S \otimes I_P + I_S \otimes H_P$ is diagonal in the computational basis, so the imaginary-time projector $e^{-\beta H/2}$ reduces to elementwise multiplication, avoiding the need for matrix diagonalization or general matrix exponentiation.

### *B. Construction of the Hamiltonian models*

#### *1) Problem Hamiltonian $\boldsymbol{H_P}$*

Each problem instance is encoded as a diagonal Hamiltonian $H_P$ of dimension $D_P = 2^{N_P}$ in the computational basis of $P$. Because the targeted problem class is classical binary optimization, only the diagonal elements of $H_P$ are populated. Consequently, $H_P$ is a diagonal, REM-like random-energy Hamiltonian (the classical, zero-transverse-field limit of the QREM discussed in Section II-G). Its levels are drawn from a uniform distribution rather than the Gaussian distribution that defines the canonical REM.

The $D_P$ diagonal entries of $H_P$ are generated in two stages:

1. Random draw.

$D_P$ values are drawn as independent and identically distributed (i.i.d.) samples from a uniform distribution $U([-R, R])$, where $R$ is a tunable spectral range parameter. In the present work, the range $R$ is varied by a parameter sweep over the values $\{0.01, 0.1, 1, 10\}$. Unless otherwise specified, $R$ is fixed at the default value of $R = 2$.

2. Desired value of degeneracy imposition.

To ensure that the constructed $H_P$ has a target GS degeneracy $g_P$, the $g_P$ lowest-energy entries of the random spectrum are identified and replaced by their common minimum value $E_0 = min(H_P^{diag})$. The remaining $D_P - g_P$ entries are left unmodified. This procedure ensures that exactly $g_P$ degenerate configurations are present at the global minimum without altering the uniform distribution statistics of the remaining higher-energy part of the spectrum. The exact-degeneracy values investigated in this work are $g_P \in \{1,4,8,12,16\}$. For the quasi-degeneracy studies of Section IV-F, a modified procedure for imposing an approximate degeneracy on $H_P$ is used. The $g_{P,Approx}$ lowest-energy entries are replaced by an arithmetic progression starting at $E_0$ and separated by a constant gap $\Delta g_P$, so the $k$-th near-degenerate state has energy $E_0 + (k - 1)\Delta g_P$ for $k = 1, \dots, g_{P,Approx}$. For different instances of $H_P$, the gap is varied over a logarithmically spaced set $\Delta g_P \in 10^{-8}, 10^{-7}, 10^{-6}, 10^{-5}, 10^{-4}$, together with $\Delta g_P = 0$, which corresponds to exact degeneracy. The energy gap between the highest near-degenerate state and the next excited state is fixed at 0.5 for all runs in which approximate degeneracy is studied.

*2) Probe Hamiltonian* $\boldsymbol{H_S}$

The probe is a small, nontrivial system satisfying the $g_S > 1$ requirement of equation (19). It comprises $N_S = 4$ qubits, giving $D_S = 16$. $H_S$ is diagonal in the computational basis, and its 16 diagonal entries are drawn from $U([-R, R])$ using the same range $R$ as that of the $H_P$. The four smallest entries are then assigned their common minimum value, fixing $g_S = 4$ throughout this work. Fixing $g_S$ in equation (19) to the same specific value isolates the dependence of $\sigma_S$ on the problem degeneracy $g_P$, which is the primary quantity of interest.

*C. Finite-temperature CTPQ state preparation*

On a quantum computer, a CTPQ state would be prepared by real-time evolution or an equivalent thermalization protocol such as VQT [40] or VarQITE [41]. One of these approaches will be necessary to apply the CTPQsd# algorithm to practical problem scales on a quantum computer with a sufficient (and currently unavailable) number of qubits. In the present numerical study, CTPQ states are instead constructed classically by imaginary-time projection of a random infinite-temperature state, following equation (11) in Sections II-D and II-E. For a single realization at inverse temperature $\beta = 1/(k_B T)$, with $k_B$ taken as unity, the procedure is:

1. Generate $| \psi_0\rangle$. Complex Gaussian coefficients $d_k = c_k' + i b_k'$ in the $D = D_S D_P$-dimensional joint Hilbert space are generated via the Box–Muller transform (equations 8–10) using independent uniform deviates $r_0^{(k)}, r_1^{(k)} \in [0,1)$. The vector is normalized to unit length.

2. Apply the imaginary-time projector. Because $\lambda = 0$ and both $H_S$ and $H_P$ are diagonal, $H$ is diagonal in the joint computational basis. Therefore, the imaginary-time projector $e^{-\beta H/2}$ can be applied via element-wise multiplication rather than general matrix exponentiation.

3. Renormalize. The projected vector is divided by $\langle \psi_0 | e^{-\beta H} | \psi_0 \rangle^{1/2}$ to obtain $| \psi_\beta \rangle$.

The diagonal structure of $H$ reduces the cost of projection from $O(D^3)$ for generic $H$ to $O(D)$ per realization [52], which makes simulation up to $N_P = 20$ ($D = 2^{24} \approx 1.7 \times 10^7$ for $N_S = 4$) tractable on a single workstation.

At the low temperatures required by the algorithm, e.g., $T$ as small as $10^{-8}$ in some cases, the imaginary-time projector $e^{-\beta E/2}$ spans a dynamic range that far exceeds the representable range of IEEE double precision, causing overflow and underflow. In this regime, Mathematica automatically promotes the projector and the associated partition sums from machine precision to arbitrary-precision arithmetic, which preserves the relative magnitudes that would otherwise be lost to overflow and keeps the projection numerically stable. The resulting low-temperature observables agree with the analytic predictions within the reported statistical uncertainties. Nevertheless, similar issues may limit the minimum temperature $T$ that can be employed across different implementations of the algorithm.

*D. Decoherence measurement and degeneracy extraction*

*1) Reduced density matrix and decoherence measure* $\boldsymbol{\sigma_S}$

For each realization $| \psi_\beta \rangle$, the reduced density matrix of the probe is obtained by partial trace over the problem subsystem $P$ following equation (3) of Section II-A. Because $H_P$ is diagonal, the partial trace reduces to $\rho_S(\beta)_{ij} = \Sigma_p a_{i,p}^* a_{j,p}$, evaluated as a sum of outer products of $D_S$-dimensional sub-vectors with no off-diagonal mixing. The decoherence measure $\sigma_S^2$ is then computed directly from the off-diagonal elements of $\rho_S$ via equation (4).

*2) Statistical averaging*

The expectation value $E(\sigma_S^2)$ appearing in equation (19) is estimated as the empirical mean over $n$ independent realizations of $| \psi_\beta \rangle$. $n = 100$ was used for all simulation runs except at $T = 10^{-8}$, where $n = 350$. Within a given simulation run (i.e., fixed $N_P$, $g_P$, $R$, $T$), $H_S$ and $H_P$ are constructed once at the start of the parameter sweep and held fixed across all $n$ realizations, but the Gaussian random vector $| \psi_0 \rangle$ is redrawn for each realization. This protocol isolates the statistical fluctuations of the CTPQ ensemble from disorder fluctuations in the Hamiltonian draw and is consistent with the canonical-typicality argument summarized in Section II-D. Convergence of the empirical mean was verified by re-running a few selected cases with $n = 1200$. Increasing $n$ improved statistical convergence at low temperatures. However, for the systems we studied, no statistically significant improvement was observed beyond $n = 350$, and hence $n > 350$ was not used for any reported data or plots.

*3) Solving for* $\boldsymbol{g_P}$

The measured $E(\sigma_S^2)$ is converted to a degeneracy estimate $g_{P,Meas}$ by solving the low-temperature limit of the perturbation result in equation (19):

$$E(\sigma^2) = (g_S - 1)/(2 g_S g_P) \times [1 - D_S D_P/((D_S D_P + 1) g_S g_P)] \tag{22}$$

This is a quadratic equation in $g_P$ with $g_S = 4$, $D_S = 16$, and $D_P = 2^{N_P}$ known by construction.

$$g_{P,Meas} = \frac{(g_S - 1)/(2g_S) \pm \sqrt{((g_S - 1)/(2g_S))^2 - 4E(\sigma^2)((g_S - 1)D_S D_P}}{2E(\sigma^2)} \tag{23}$$

The two roots are computed analytically from the averaged $E(\sigma_S^2)$ of all samples. The physically admissible root, satisfying $1 \le g_{P,Meas} \le D_P$, is selected. For exact-degeneracy cases, $g_{P,Meas}$ is reported as a real number with its statistical error. For the quasi-degeneracy studies, the same methodology is applied, and the resulting $g_{P,Meas}$ is interpreted as the count of the approximate-degenerate minima $g_{P,Approx}$ when the operating temperature lies in the window $T_2 < T < T_1$ identified in Section IV-F.

*E. Implementation, software, and computational resources*

All simulations were implemented in Mathematica and run on a single machine (Intel Core i7-12700H, 14 physical cores / 20 threads, 64 GB RAM). Extended-precision evaluation relied on Mathematica's built-in arbitrary-precision arithmetic, invoked automatically in the low-temperature regime. The random-number generator was seeded (SeedRandom[1234]) at the beginning of each run for reproducibility. The largest system simulated ($N_S = 4$, $N_P = 20$) required a complex-valued state vector of $2^{24}$ entries ($\approx 256MB$) in machine precision, so memory was not a binding constraint. The dominant computational expense arises from the time required to perform large ensembles of up to 350 realizations per configuration with arbitrary-precision arithmetic in the low-temperature regime.

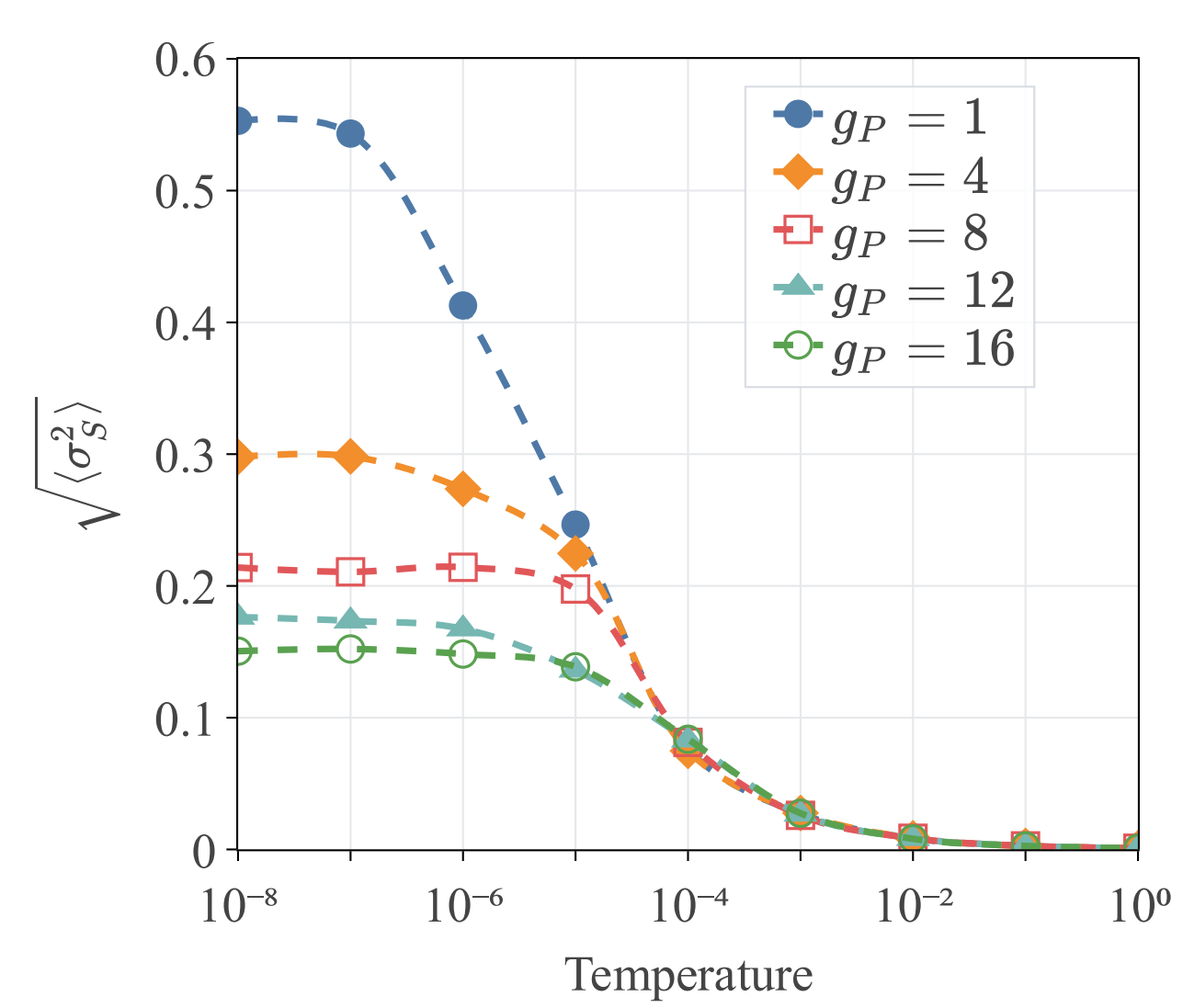

Fig. 1. The temperature dependence of the degree of decoherence $\sigma_S$ of the Probe system $S$, shown for a few different values of the degeneracy of the problem $g_P$. Curves from the top to bottom correspond to $g_P$ =1, 4, 8, 12, and 16, for the case when $N_P = 20$ (the largest $N_P$ investigated in this work). The low-temperature region of constant $\sigma_S$, corresponding to when the state of the entirety $S + P$ approaches the ground state, is used to calculate $g_{P,Meas}$ in this work. In this and other figures, the dashed lines are guides to the eye.

## IV. Results

### A. Temperature dependence of decoherence and the low-temperature regime

As discussed in Section II-F, equation (19) relates the degeneracy $g_P$ of the problem $P$ with the decoherence $\sigma_S$ of the small probe system $S$ at sufficiently low temperatures. We therefore begin by characterizing the temperature dependence of $\sigma_S$ to identify the temperature window in which the CTPQsd# algorithm can be applied to reliably predict the degeneracy of the problem $P$. Fig. 1 plots the temperature dependence of the degree of decoherence $\sigma_S$ of $S$ for the case when $N_P = 20$, which is the largest $N_P$ that was investigated in this work. At lower temperatures, higher numerical precision is required to compensate for numerical underflow and overflow and calculate the decoherence values. This precision requirement, together with the growth of the quantum Hilbert space, increases the time and memory costs of classical simulation, particularly in the low-temperature regime where the CTPQsd# algorithm is effective for the Hamiltonians studied here. These factors limit the number of qubits that can be realistically simulated on classical computers.

The five curves in Fig. 1 correspond to five different values of the problem's degeneracy, $g_P$. From top to bottom, the curves correspond to $g_P$ = 1, 4, 8, 12, and 16. The $H_S$ with $N_S = 4$ and $g_S = 4$ is kept fixed for all runs while varying $H_P$. In accordance with equation (17), $\sigma_S$ increases with a decrease in temperature $T$ and saturates at sufficiently low $T$, which corresponds to the regime where the entirety $S + P$ approaches the ground state manifold, i.e., the subspace spanned by the ground state(s). This low-temperature region is used to extract $g_{P,Meas}$ in this work. A key observation, evident in Fig. 1, is that the onset temperature for saturation of $\sigma_S$ is relatively independent of $g_P$ (see also the discussion for Fig. 5 below). This threshold temperature is denoted by $T_1$ in this paper.

### B. Scaling of the saturation temperature with problem size

Having established $T_1$ as the operating threshold for the CTPQsd# algorithm, which may depend on other parameters of the algorithm, the dependence of $T_1$ on the dimensionality of the problem is investigated next. Fig. 2 presents the temperature dependence of $\sigma_S$ for various values of $N_P$ at two fixed degeneracies: $g_P = 1$ (Fig. 2a) and $g_P = 16$ (Fig. 2b). Curves from the top to the bottom correspond to $N_P$ = 8, 12, 16, and 20, and demonstrate that increasing the number of problem qubits $N_P$ requires progressively lower temperatures $T$ to reach the low-temperature regime $T < T_1$, where $\sigma_S$ saturates. Therefore, special care must be taken to ensure the low-temperature regime is properly accessed, particularly for high-dimensional problems.

With $2^{N_P}$ energy levels drawn from a fixed range $R$, the characteristic gap between the GS manifold and the nearest excited level for $2^{N_P}$ problem configurations scales for uniform distribution as $\Delta E_P \sim R \cdot 2^{-N_P}$. The low-temperature limit of the perturbative relation eq. (19), the one the algorithm uses, is valid only when the thermal state is essentially confined to the GS manifold. In this state, the occupation of the nearest excited level is expected to be negligible, i.e. $e^{(-\Delta E_P/k_B T_1)} \ll 1$. Hence, thermally resolving the GS manifold requires $k_B T_1$ of order $\Delta E_P$ or below. The operating threshold is therefore expected to scale as $T_1 \propto R \cdot 2^{-N_P}$, consistent with the dependences on $N_P$ and $R$ observed in Figs. 2, 4, 6, and 7.

### C. Accuracy of the CTPQsd# algorithm

Next, the performance of the CTPQsd# algorithm was evaluated for the two largest values of $N_P$ investigated in this work. Fig. 3 plots the calculated degeneracy $g_{P,Meas}$ determined using equation (19), against the actual degeneracy $g_P$ for five problem systems $P$ with different degeneracy values. The two curves are for $N_P = 20$ and $N_P = 16$. All the values of $g_{P,Meas}$ were extracted from $\sigma_S$ at a sufficiently low temperature $T = 10^{-8}$, well within the regime identified in Figs. 1-2. For each data point, averages for $\sigma_S^2$ were computed for 350 independent realizations of the low-temperature CTPQ states $|\psi_\beta\rangle$. The error bars show the standard error of the mean (SEM) for $N_P = 20$, computed by dividing the standard deviation of $\sigma^2$ across the 350 realizations by $\sqrt{n}$ (n = 350) and propagating the result through equation (23). Because the reported $g_{P,Meas}$ is extracted from the ensemble-averaged $\sigma^2$, we use SEM, rather than the standard deviation, as the measure of the uncertainty of the plotted estimate, and it decreases as $1/\sqrt{n}$ with the number of realizations. As shown in Fig. 3, $g_{P,Meas}$ remains in close agreement with the degeneracy $g_P$ across all simulated scenarios. While individual CTPQ realizations exhibit significant variance, i.e., the single-shot standard deviation is substantially larger than the SEM, the statistical mean $g_{P,Meas}$ converges reliably to the true value, with an uncertainty (SEM) that can be made arbitrarily small by increasing the number of realizations. Consequently, the ambiguity inherent in single-shot measurements can be systematically eliminated by averaging over a sufficiently large ensemble of CTPQ states, allowing for identification of correct degeneracy.

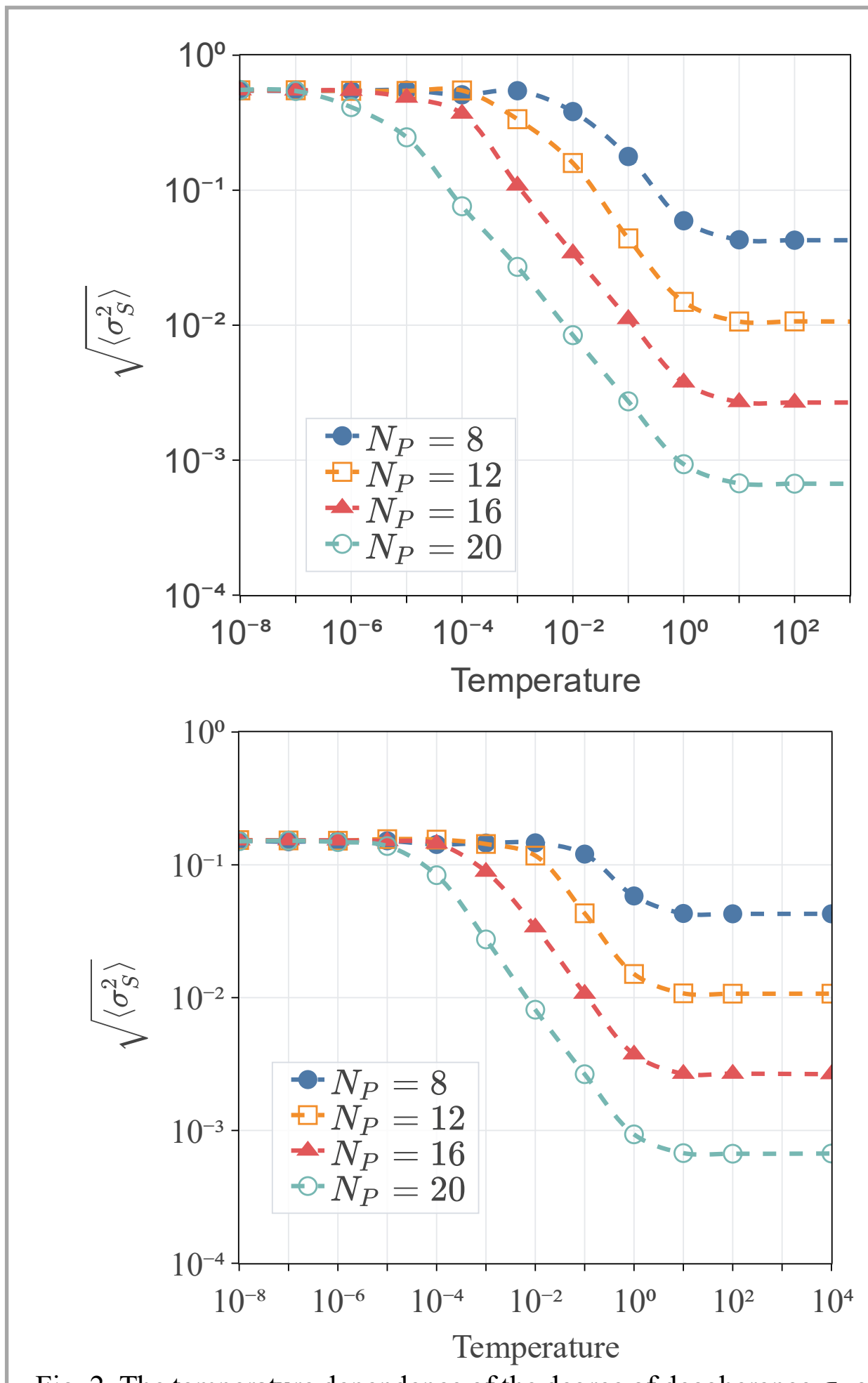

Fig. 2. The temperature dependence of the degree of decoherence $\sigma_S$ of the Probe $S$, shown for a few different values of the $N_P$: (a) for $g_P = 1$ and (b) for $g_P = 16$. Curves from the top to bottom correspond to $N_P = 8$, 12, 16 and 20. An increasing number of qubits $N_P$ requires lower $T$ for reaching the required low-temperature region of constant $\sigma_S$.

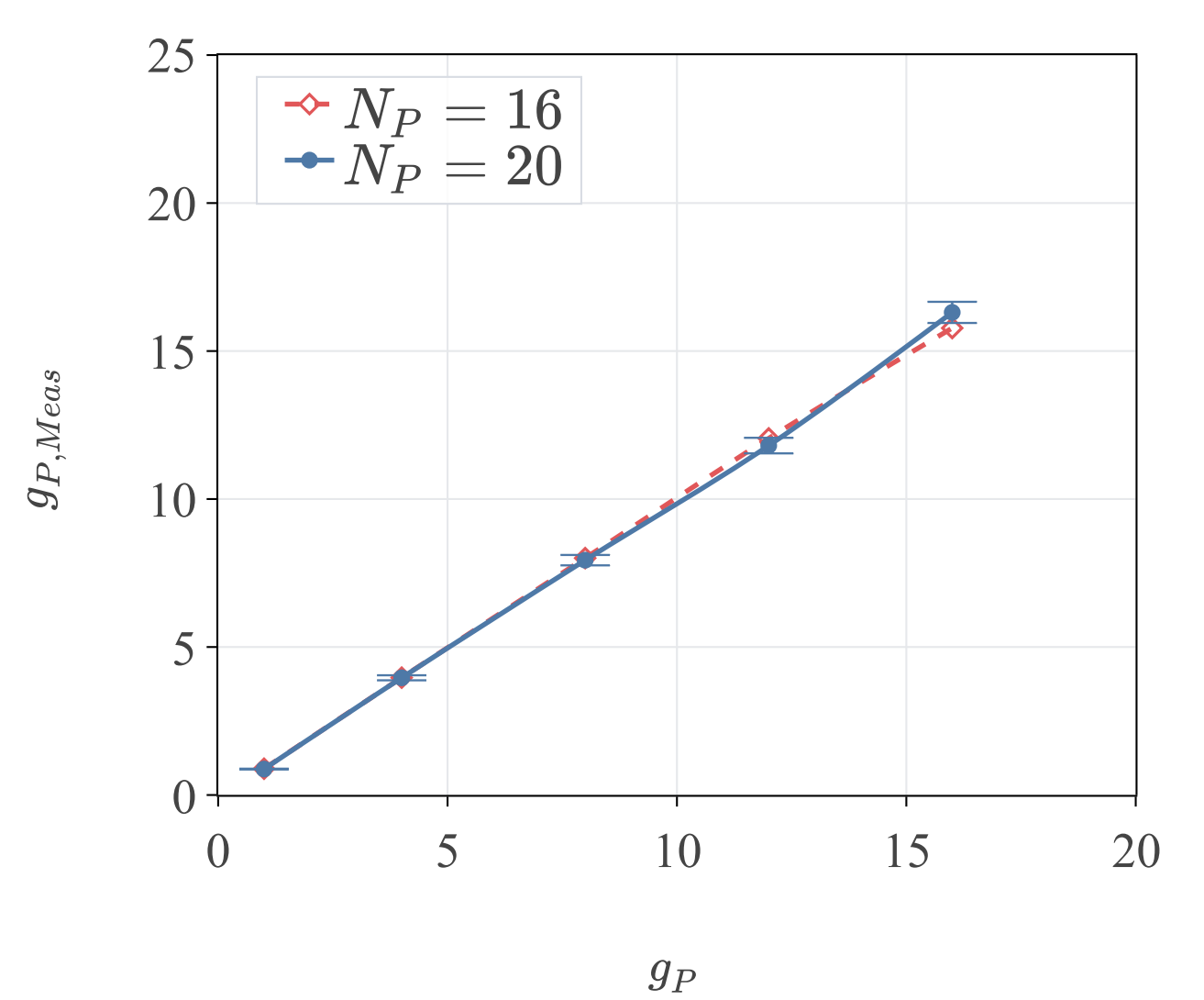

Fig. 3. The value of the degeneracy of the problem $g_{P,Meas}$ determined using CTPQsd# versus the actual degeneracy $g_P$. The two curves are for the two cases of the biggest numbers of qubits in $P$ investigated in this work: $N_P = 20$ and $N_P = 16$. $g_P$ was extracted from the value of the decoherence at $T = 10^{-8}$. For both $N_P = 16$ and $N_P = 20$, 350 averages with different random $\Psi_0$ were used. The error bars show the standard error of the mean (SEM) over the 350 realizations for $N_P = 20$. Although individual single-shot realizations exhibit a much larger spread, the ensemble mean $g_{P,Meas}$ is determined precisely, as reflected by the small SEM, and aligns closely with the true values.

Owing to the time and memory limitations of classically simulating quantum states, it is not possible to investigate larger problems ($N_P > 20$) and assess whether substantially larger numbers of qubits in $P$ would increase the error in predicting $g_P$. If this issue is encountered in practical realizations of the CTPQsd# algorithm, the precision of the algorithm can be improved by increasing the number of samples (i.e., preparations of the CTPQ state) used to calculate $g_{P,Meas}$. Moreover, canonical typicality, discussed in Section II-D, suggests that for a sufficiently large entirety $S + P$, a relatively small number of CTPQ states is expected to suffice for reliable estimation of equilibrium expectation values of observables, including $\sigma_S^2$ used in the CTPQsd# algorithm.

### *D. Relative error of degeneracy determination at non-optimal temperatures*

To provide quantitative insight into the impact of non-optimal temperatures $T$ on the degeneracy determination by the CTPQsd# algorithm, the temperature dependence of the relative error $|g_{P,Meas} - g_P|/g_P$ in determining the degeneracy $g_{P,Meas}$ was investigated in Figs. 4 and 5. Fig. 4 shows $|g_{P,Meas} - g_P|/g_P$ as a function of temperature for several values of $N_P$, for the case $g_P = 16$. Curves from the bottom to top correspond to the growing number of qubits in the problem, i.e., $N_P = 8$, 12, 16, and 20. Confirming the qualitative observations from Fig. 2, the results of Fig. 4 indicate that problems with a larger number of qubits $N_P$ require lower temperatures, i.e., a lower $T_1$ threshold for a correct determination of $g_P$, all else being equal.

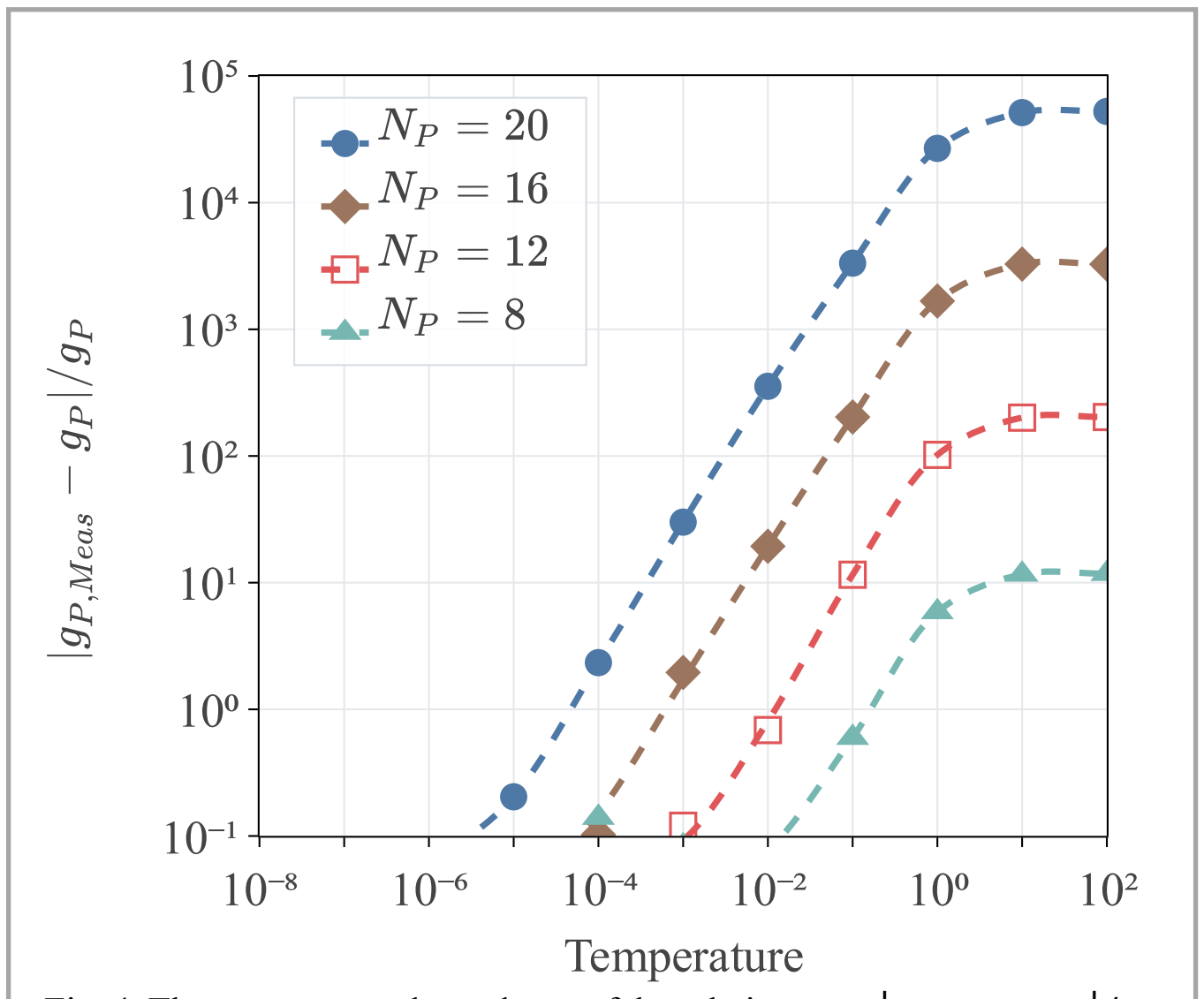

Fig. 4. The temperature dependence of the relative error $|g_{P,Meas} - g_P|/g_P$ in determining the degeneracy $g_{P,Meas}$, shown for a few different values of the number of qubits in the problem $N_P$. The actual degeneracy in this figure $g_P$ is 16. Curves from the bottom to top correspond to the growing number of qubits in the problem: $N_P$ = 8, 12, 16 and 20. The results indicate that an increasing number of qubits $N_P$ requires lower $T$ for a correct determination of $g_P$ when other conditions are similar.

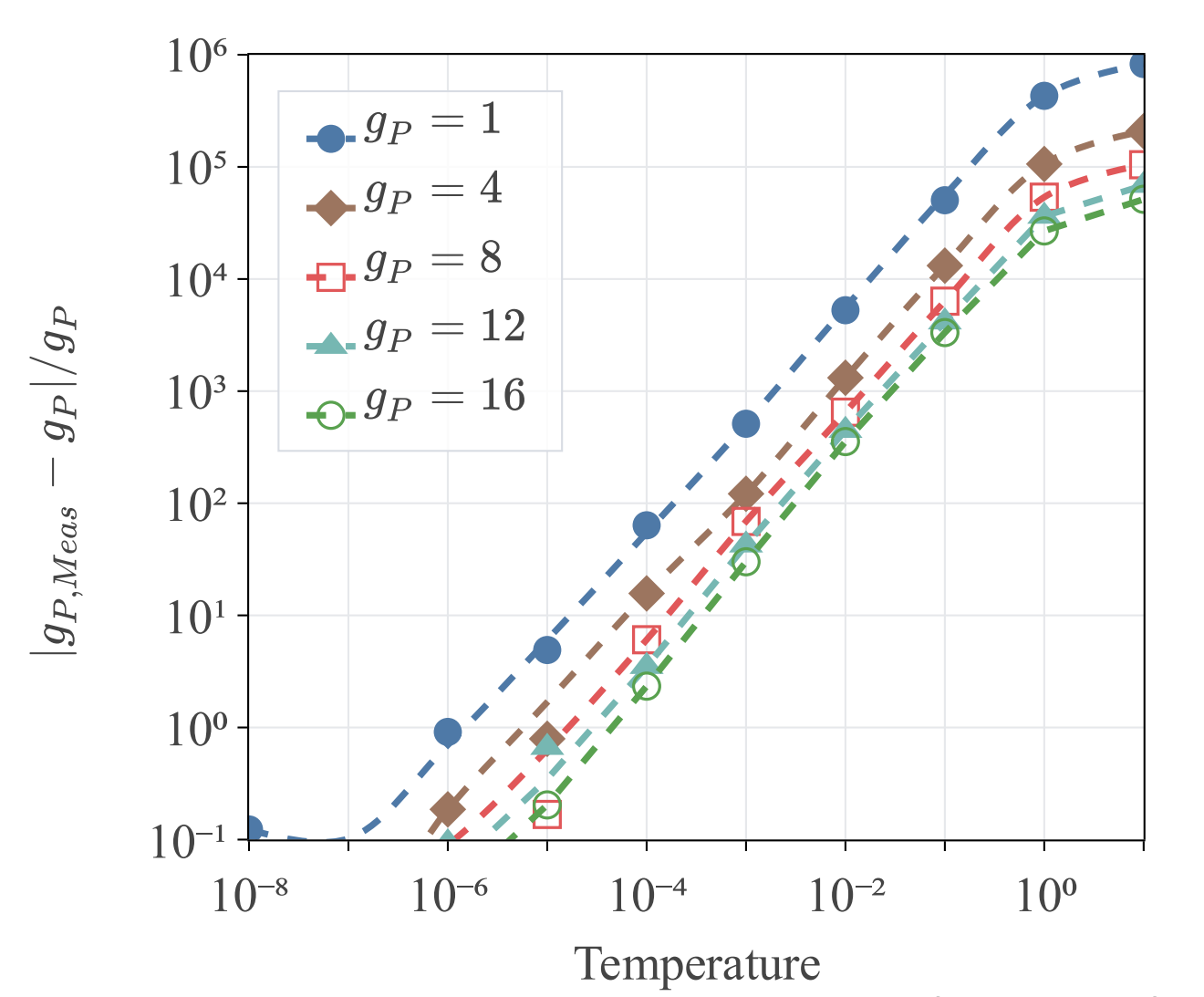

Fig. 5. The temperature dependence of the relative error $|g_{P,Meas} - g_P|/g_P$ in determining the degeneracy $g_{P,Meas}$ for a few different values of the degeneracy $g_P$. All the results in this figure are for $N_P$ = 20. Curves from the top to bottom correspond to $g_P$ = 1, 4, 8, 12 and 16. We observe that problems with lower $g_P$ may require somewhat lower $T$ for a correct determination of $g_P$ when other conditions are similar.

Fig. 5 examines the complementary question of whether the low-temperature regime $T < T_1$ depends on the problem degeneracy $g_P$. Fig. 5 plots $|\ g_{P,Meas} - g_P\ |/g_P$ as a function of temperature for several degeneracy values $g_P$=1, 4, 8, 12, and 16, for fixed problem size $N_P = 20$. In contrast to the strong dependence on $N_P$ reported in Fig. 4, the threshold temperature $T_1$ required for accurate degeneracy extraction shows only a weak dependence on $g_P$. Problems with lower $g_P$ may require somewhat lower $T$ for a correct determination of $g_P$ when other conditions are similar.

### *E. CTPQsd# algorithm's dependence on the diagonal energy range parameter*

We next investigate the sensitivity of the CTPQsd# algorithm, not only to the problem dimensions $N_P$ and problem degeneracy $g_P$ but also to other characteristics of the encoding Hamiltonians $H_S$ and $H_P$. As discussed in Sections I and II, the use of a diagonal REM-like Hamiltonian in this work represents an initial step towards establishing a proof of concept for the applicability of the CTPQsd# algorithm to a nearly arbitrary classical problem. One of the parameters of a REM-like Hamiltonian, which is expected to have practical importance for the execution of the CTPQsd# algorithm, is the range of values of the randomly generated diagonal elements of $H_S$ and $H_P$. Fig. 6 presents the temperature dependence of $\sigma_S$ for different values of the range $R$ of the diagonal elements of Hamiltonians $H_S$ and $H_P$. The results are for the case of $N_P = 20$, the largest $N_P$ investigated in this work, and $g_P = 8$, with curves from bottom to top corresponding to the increasing values of range $R$ = 0.01, 0.1, 1, and 10. The figure suggests that Hamiltonians with lower spectral variance require a lower threshold temperature, $T_1$, to enter the required low-temperature regime of constant $\sigma_S$ .

For a more quantitative evaluation of the effect of $R$ of the Hamiltonian on the temperature range required for the correct prediction of degeneracy $g_{P,Meas}$ from the CTPQsd# algorithm, the temperature dependence of the relative error $|\ g_{P,Meas} - g_P\ |/g_P$ was evaluated for a few different values of $R$ in Fig. 7. Curves from the bottom to the top correspond to the decreasing range $R$ = 10, 1, 0.1, and 0.01, with $N_P = 20$, and $g_P = 8$. The figure shows that a narrower range $R$ of the diagonal values of Hamiltonians requires lower $T$ for an accurate determination of $g_P$ (i.e., a lower threshold temperature $T_1$).

Given the relative arbitrariness of the REM-like Hamiltonians used here, the results shown in Figs. 6 and 7 suggest that problem Hamiltonians with smaller spectral variance will generally require lower operating temperatures for the CTPQsd# algorithm to predict the degeneracy $g_P$ accurately, and this trend is likely to extend to other classes of problem Hamiltonians.

### *F. Determination of quasi-degeneracy*

The results presented thus far pertain to finding the exact degeneracy, defined as the number of local minima with energies exactly equal to that of the global minimum (i.e., the GS). The remainder of the simulations in this work focus on determining the quasi-degeneracy or approximate degeneracy $g_{P,Approx}$, defined as the number of local extrema whose energies fall within a small tolerance of the global optimum.

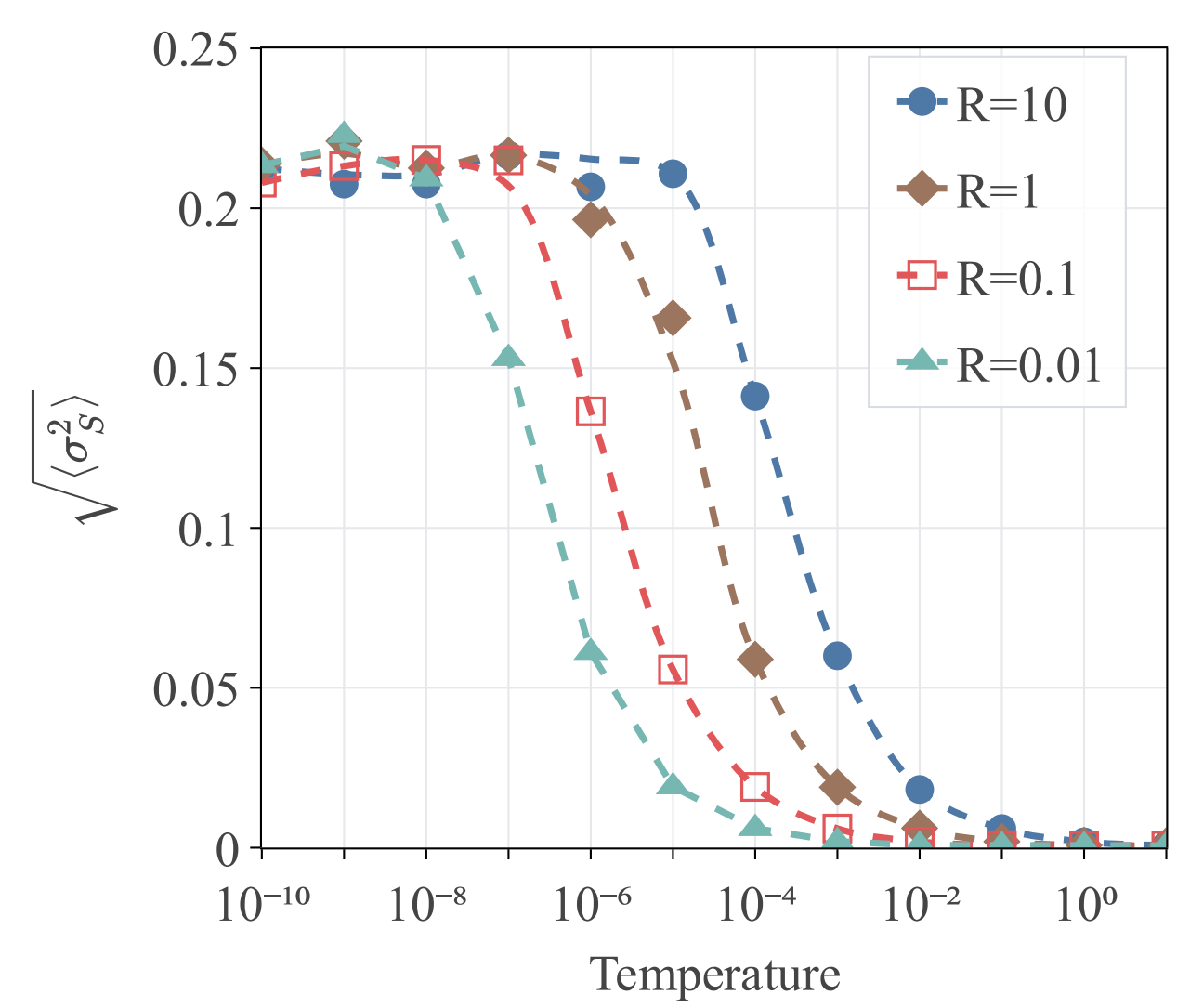

Fig. 6. The temperature dependence of the degree of decoherence $\sigma_S$ of the probe $S$, shown for a few different values of the range of the Hamiltonian $H_P$. Curves from the bottom to top correspond to the increasing range $R =$ 0.01, 0.1, 1 and 10, for the case when $N_P = 20$ and $g_P = 8$. The results indicate that lower spectral range requires lower $T$ to reach the required low-temperature region of constant $\sigma_S$.

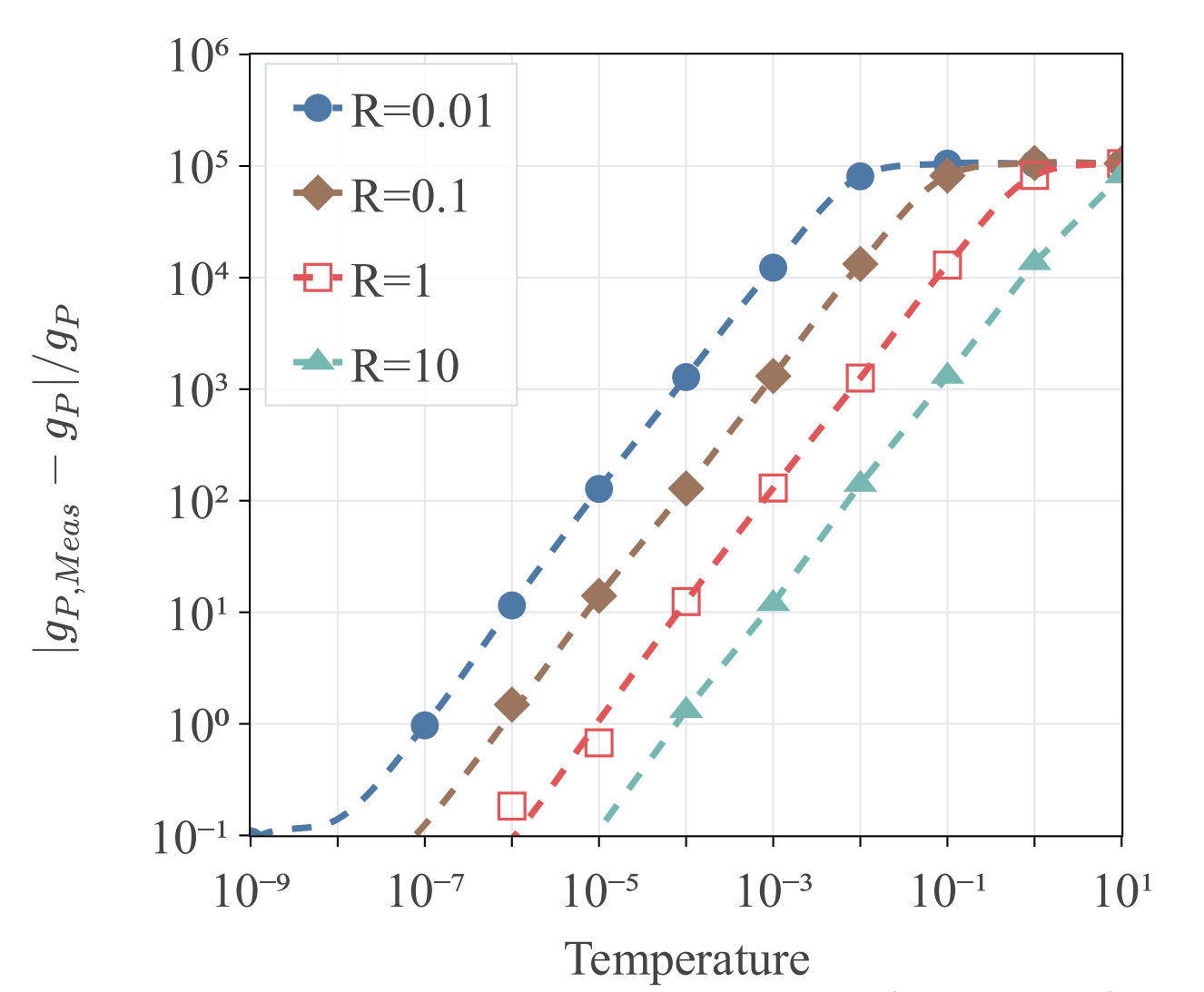

Fig. 7. The temperature dependence of the relative error $|g_{P,Meas} - g_P|/g_P$ in determining the degeneracy $g_{P,Meas}$, shown for a few different values of the range of the Hamiltonian $H_P$. Curves from the bottom to top correspond to the decreasing range, $R = 10$, 1, 0.1 and 0.01, for the case when $N_P =$ 20 and $g_P = 8$. As follows from the figure, lower spectral range requires lower $T$ for a correct determination of $g_P$ when other conditions are similar (see also Fig. 6 for the relevant dependence of $\sigma_S$).

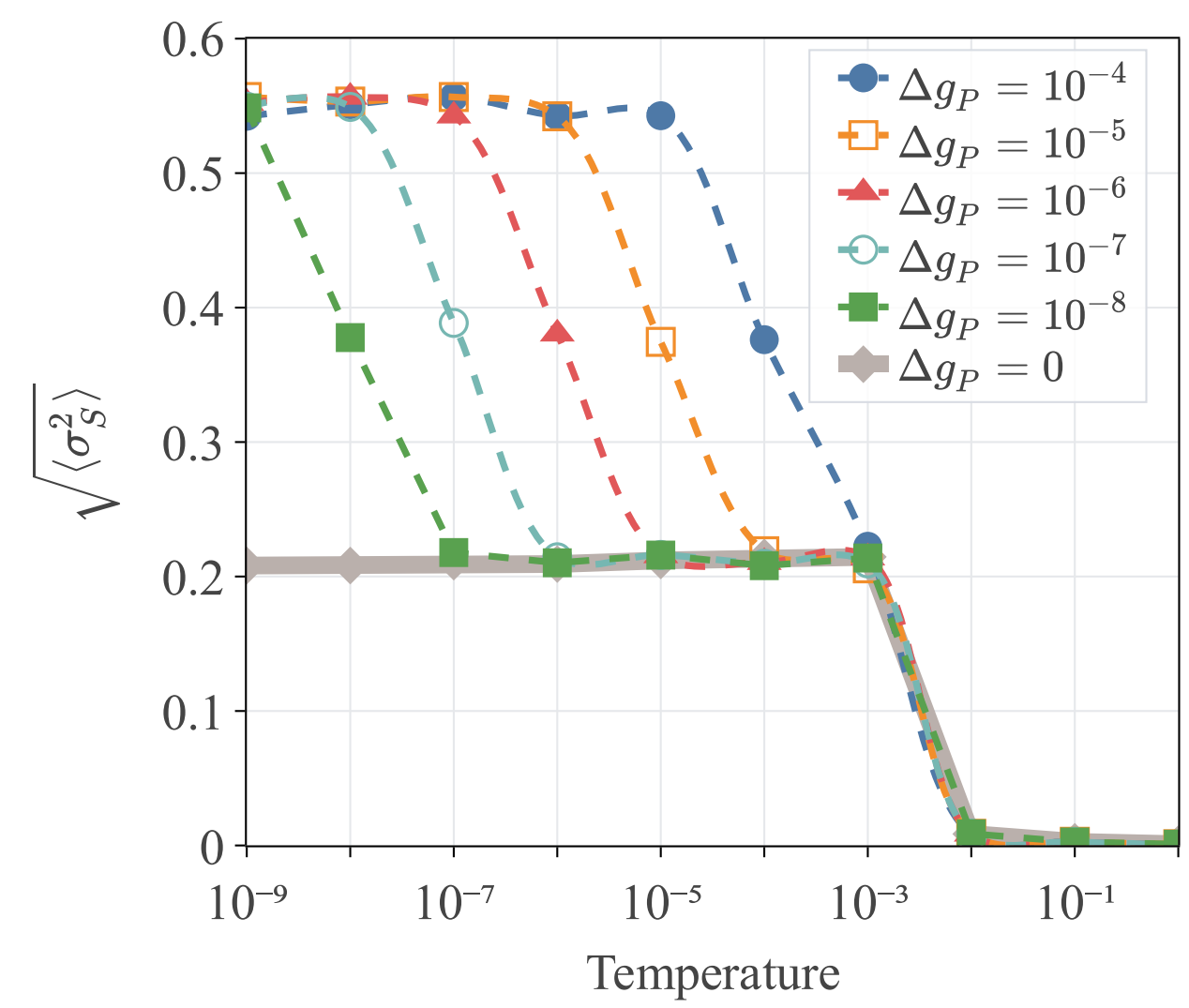

Fig. 8. The temperature dependence of the degree of decoherence $\sigma_S$ of the probe $S$ for the approximate degeneracy case with $g_{P,Approx} = 8$ and $N_P =$ 20. Unlike Figs. 1-7, the eight lowest energy (nearly-degenerate) states do not have the same energy but differ by a small energy gap $\Delta g_P$. Curves from the bottom to top correspond to the increasing energy gap, $\Delta g_P = 0$, $10^{-8}$, $10^{-7}$, $10^{-6}$, $10^{-5}$, and $10^{-4}$. The bottom-most light-gray thicker curve corresponds to the exact-degeneracy case (i.e., $\Delta g_P = 0$). When $T$ is reduced below a certain threshold, the low-temperature region used to calculate $g_{P,Meas}$ is different from the case of the exact degeneracy and should give a different value of $g_{P,Meas}$. As could be expected, for large enough gaps, the temperature window to determine the approximate degeneracy (the first lower plateau) disappears entirely.

Fig. 8 shows the temperature dependence of $\sigma_S$ for the case of the approximate degeneracy with $g_{P,Approx} = 8$ and $N_P =$ 20. Unlike in Figs. 1-7, where the lowest-energy states all have the same energy, here the energies of the eight lowest-energy states differ from each other by a small constant energy gap $\Delta g_P$. Curves from the bottom to top correspond to an increasing energy gap $\Delta g_P = 0$, $10^{-8}$, $10^{-7}$, $10^{-6}$, $10^{-5}$, and $10^{-4}$, with the bottom-most curve $\Delta g_P = 0$ representing the exact-degeneracy case. When comparing the $\sigma_S$-versus-$T$ curves for the exact and the approximate degeneracy cases, the latter exhibits an additional threshold at a lower temperature $T_2 < T_1$, where $T_2$ marks the saturation threshold of $\sigma_S$ for the exact degeneracy, and $T_1$, discussed earlier, now becomes the onset of the regime to detect approximate degeneracy. As $T$ is reduced below $T_2$, the calculated value of $g_{P,Meas}$ would be the number of exact-degenerate states, which would be lower than the number of approximate-degenerate states $g_{P,Approx}$ calculated at $T_2 < T < T_1$.

Fig. 9 shows the temperature dependence of the absolute error $|\, g_{P,Meas} - g_P \,|$ in determining the degeneracy $g_{P,Meas}$, for the same set of approximate-degenerate values as in Fig. 8. In the region of low temperature $T < T_1$ used in calculating $g_{P,Approx}$ in Fig. 8, further reduction of $T$ sufficiently below the threshold $T_1$ causes incremental exclusion of the higher-energy states in the group of nearly-degenerate states from contributing to the determined $g_{P,Meas}$. For example, in Fig. 9, for $\Delta g_P = 10^{-5}$, the range of $T$ from ~0.001 to ~0.0001 allows the determination of the count of approximate-degenerate states $g_{P,Approx}$ (i.e., the error $|\, g_{P,Meas} - g_P \,|$ in Fig. 9 in this temperature range is close to zero). At temperatures $T < T_2 = 0.0001$, $g_{P,Meas}$ begins decreasing again (which corresponds to the increasing value of the error) and ultimately saturates at the count of exact-degeneracy, which in this case is $g_{P,Meas} = 1$, irrespective of the number of approximate-degenerate states. Smaller values of $\Delta g_P$ broaden the temperature window in which the approximate degeneracy is reflected in the count of degeneracy $g_{P,Meas}$. In contrast, at too large a $\Delta g_P$, no temperature range exists that allows for a reliable determination of $g_{P,Approx}$.

This two-threshold structure has significant practical implications. Given a user-specified energy tolerance $\Delta g_{P,Approx}$ defining what counts as a "good-enough" extremum of the cost function, the CTPQsd# algorithm could be applied at a temperature satisfying $T_2 < T < T_1$ to count all such approximate-degenerate states.

## V. Conclusion

This work investigated a decoherence-based quantum algorithm, CTPQsd#, for determining the GS degeneracy $g_P$ of a broad class of arbitrary classical optimization problems without enumerating their solutions. Building on the perturbative relation derived by Novotny et al.[36,37], this work (i) demonstrated for the first time that the algorithm is capable of successfully determining degeneracy for a wide range of problem settings, (ii) applied the underlying perturbative relationship for the degeneracy, previously considered only for spin-1/2 Heisenberg Hamiltonians, to the structurally distinct, diagonal REM-like Hamiltonian with no transverse field that encodes classical binary optimization problems, and (iii) systematically characterized the parameters that govern the accuracy to extract a quantitative estimate of the degeneracy $g_{P,Meas}$. We emphasize that, although the present study is restricted to classical (diagonal) problem Hamiltonians, the success on REM-like Hamiltonians, structurally unrelated to the Heisenberg models of prior work, indicates that the relation is not an artifact of those models and that the method should extend to broader problem classes.

Across all simulated instances, with a fixed four-qubit probe ($g_S = 4$) and problem sizes up to $N_P = 20$ (i.e., the largest number of qubits that we could handle in our simulations), the recovered $g_{P,Meas}$ agreed with the true degeneracy for every value tested ($g_P \in 1,4,8,12,16$) at sufficiently low temperature. The accuracy of the method is controlled by a threshold temperature $T_1$, below which $\sigma_S$ becomes degeneracy-sensitive. We found that $T_1$ decreases with problem size $N_P$, is not significantly sensitive to problem degeneracy $g_P$, and is lower for Hamiltonians of smaller spectral range $R$, which means that problems with a smaller energy spread require correspondingly lower operating temperatures. For approximate-degenerate rather than exactly degenerate minima, the algorithm exhibits a two-threshold structure. Within the window $T_2 < T < T_1$, it counts the approximate degeneracy $g_{P,Approx}$, i.e., the number of local minima lying within a tolerance $\Delta g_P$ of the global minimum, while below $T_2$, it recovers the exact degeneracy count of those states, providing a tunable, tolerance-aware route to counting approximate-degenerate solutions. Because the measurement is performed only on the probe, the cost of the quantum state tomography is determined by the probe dimension rather than the dimension of the exponentially large problem Hilbert space, directly addressing the measurement bottleneck that constrains many near-term counting algorithms.

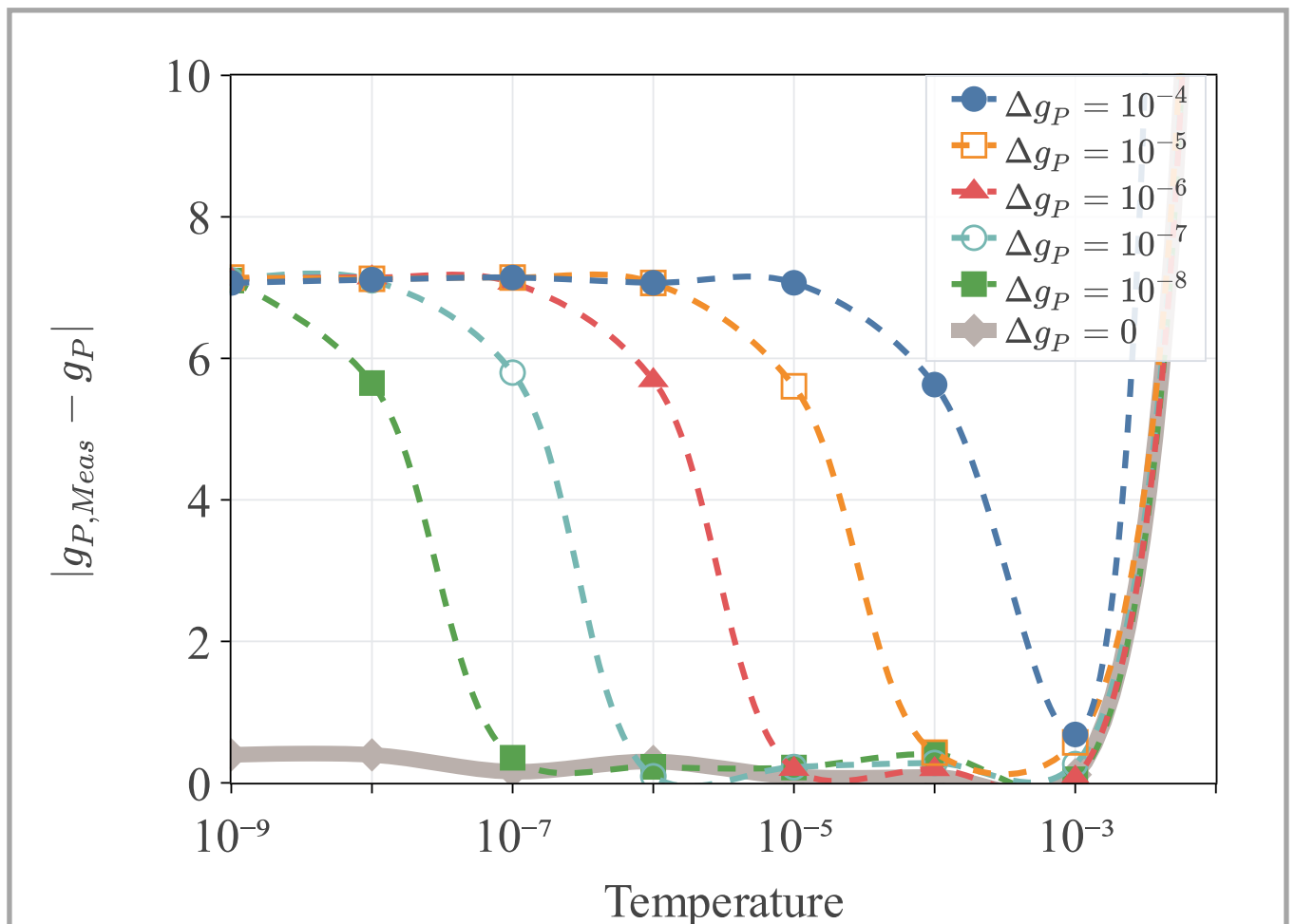

Fig. 9. The temperature dependence of the absolute error $|g_{P,Meas} - g_P|$ in determining the degeneracy $g_{P,Meas}$, for the case of the "approximate degeneracy", with $g_{P,Approx} = 8$, but the eight lowest energy (nearly-degenerate) states do not have the same energy (as in Figs. 1-7) but differ by a small energy gap $\Delta g_P$. Curves from the bottom to top correspond to the increasing energy gap: $\Delta g_P = 0$, $10^{-8}$, $10^{-7}$, $10^{-6}$, $10^{-5}$, and $10^{-4}$. The light-gray thicker curve corresponds to the exact-degeneracy case (i.e., $\Delta g_P = 0$). For all the curves, $N_P = 20$. The figure shows that a two-threshold regime is observed, in which an intermediate temperature allows counting near-degenerate minima.

When running the algorithm on real quantum computing hardware, there may be challenges with problem sizes much larger than those simulated classically in this paper. One concern is that reliably estimating $E(\sigma_S^2)$ might require a larger number of independent CTPQ preparations. This concern, however, is plausibly offset by considerations of canonical typicality, which predicts that for a sufficiently large entirety $S + P$, a comparatively small number of CTPQ states is expected to reproduce equilibrium expectation values of observables to within small fluctuations. Whether this favorable scaling holds in practice across the relevant temperature and size ranges is an important question for future work.

The following significant challenges remain prior to the quantum hardware realization of the algorithm, but none are specific to the CTPQsd# algorithm. First, the problem Hamiltonian must be encoded, and its dynamics implemented using the available gate sets. For diagonal cost Hamiltonians in classical optimization, this can be achieved by applying Pauli-Z rotations in the cost layers of QAOA-type circuits. Second, a CTPQ state must be prepared in the required low-temperature regime $T < T_1$ with finite circuit depth. Potential approaches include variational thermalization schemes, such as VQT and VarQITE, which warrant further exploration. Third, the reduced density matrix of the probe must be reconstructed by quantum state tomography to evaluate $\sigma_S^2$ from its off-diagonal elements. Because the probe is only four qubits, this readout is relatively inexpensive and largely insulated from the multiplicative accumulation of readout errors that plagues full-register measurements. The probe density matrix is $16 \times 16$ independent of $N_P$, so the tomography overhead is fixed, whereas tomography over the full problem register would scale with its $2^{N_P}$-dimensional state space ($\approx 10^6$ dimensions already at $N_P$ = 20). Beyond these, device noise must be controlled through error mitigation, as in any near-term protocol. Future studies must assess how readout noise and finite sampling affect $g_{P,Meas}$.

With the quantum information and quantum computing communities actively addressing each of these problems for the benefit of many algorithms, and with quantum hardware maturing rapidly, the algorithm developed in this work is a promising direction for a scalable solution for the important task of determining the number of global extrema or nearly optimal local extrema of an arbitrary classical problem.

## Acknowledgment

Y K acknowledges support from the U.S. National Science Foundation under Grant No. CCF-2211841. M N acknowledges support from U.S. Department of Energy under Grant No. DE-SC0024286. Any opinions, findings, conclusions, or recommendations expressed in this material are those of the authors and do not necessarily reflect the views of the U.S. National Science Foundation or the U.S. Department of Energy.

## Data availability statement

The data for this paper are available at: https://github.com/MalayMDas/DegeneracyCountingQuantumAlgorithm